\documentclass[letterpaper, 10 pt, conference]{ieeeconf}  
\usepackage{xcolor}

\IEEEoverridecommandlockouts                              

\usepackage{algorithm}
\usepackage{algorithmic}
\usepackage[bookmarks=true]{hyperref}
\usepackage{hyperref}
\usepackage{biblatex}
\usepackage{amsmath} 
\usepackage{amssymb}  
\usepackage{booktabs}
\usepackage{multirow}
\usepackage{multicol}
\usepackage{graphicx}
\usepackage[T1]{fontenc}

\usepackage{colortbl}
\usepackage{makecell}
\usepackage{bm}

\title{\LARGE \bf
SIME: Enhancing Policy \underline{S}elf-\underline{I}mprovement\\with \underline{M}odal-level \underline{E}xploration
}

\author{Yang Jin$^{*1}$, Jun Lv$^{*1,3}$,  Wenye Yu$^{1}$, Hongjie Fang$^{1}$, Yong-Lu Li$^{1,2}$, Cewu Lu$^{1,2,3}$
\thanks{*Equal Contribution}
\thanks{$^{1}$Shanghai Jiao Tong University, Shanghai, CHINA}%
\thanks{$^{2}$Shanghai Innovation Institution, Shanghai, CHINA}%
\thanks{$^{3}$Noematrix Ltd., Shanghai, CHINA}}

\begin{document}

\maketitle
\thispagestyle{empty}
\pagestyle{empty}

\begin{abstract}

Self-improvement requires robotic systems to initially learn from human-provided data and then gradually enhance their capabilities through interaction with the environment. This is similar to how humans improve their skills through continuous practice. However, achieving effective self-improvement is challenging, primarily because robots tend to repeat their existing abilities during interactions, often failing to generate new, valuable data for learning. In this paper, we identify the key to successful self-improvement: modal-level exploration and data selection. By incorporating a modal-level exploration mechanism during policy execution, the robot can produce more diverse and multi-modal interactions. At the same time, we select the most valuable trials and high-quality segments from these interactions for learning. We successfully demonstrate effective robot self-improvement on both simulation benchmarks and real-world experiments. The capability for self-improvement will enable us to develop more robust and high-success-rate robotic control strategies at a lower cost. Our code and experiment scripts are available at \href{https://ericjin2002.github.io/SIME/}{ericjin2002.github.io/SIME}.

\end{abstract}

\section{Introduction}

In recent years, data-driven robot learning, particularly imitation learning~\cite{brohan2023rt, chi2024diffusionpolicy,florence2022implicit,mandlekar2021matters, shafiullah2022behavior}, has achieved remarkable progress, demonstrating impressive capabilities in various manipulation tasks. The key driving force behind this advancement is the availability of large-scale, high-quality datasets~\cite{fang2024rh20t, droid, open_x_embodiment_rt_x_2023}, which allow policies to generalize effectively across diverse scenarios. However, despite these advancements, imitation learning still faces critical limitations that hinder its scalability and broader applicability.

One such challenge is the high cost of data collection. Unlike vision or language domains~\cite{achiam2023gpt, kirillov2023segment}, where large datasets can be collected from existing sources, robot learning requires active interaction with the environment to acquire meaningful data, where expert demonstrations are recorded using hardware platforms~\cite{chi2024universal, fang2025airexo, aloha}, making data collection labor-intensive and expensive. Another fundamental limitation is that since the policy is trained to mimic expert behavior, its performance is inherently bounded by the quality and diversity of the demonstrations in the dataset. Without mechanisms for further improvement, the robot typically cannot surpass the abilities of the demonstrators. This issue becomes particularly pronounced when demonstrations are suboptimal~\cite{chen2024towards} or fail to cover the full distribution of task variations.

To address these challenges, we explore an alternative paradigm: enabling robots to improve themselves by learning from their own interaction experiences during deployment. The goal is to reduce dependence on human-collected data while allowing policies to refine their skills beyond the limitations of initial demonstrations. Existing approaches to this problem typically rely on reinforcement learning~(RL)~\cite{haldar2023teach,luo2024serl,yuan2024policy} or human-in-the-loop intervention~\cite{jiang2024transic,liu2022robot,luo2024precise,luo2024human}. However, RL suffers from poor sample efficiency. Meanwhile, human-in-the-loop methods still demand significant human effort, limiting their practicality for large-scale deployment.

\begin{figure}[t]
    \vspace{2mm}
    \centering
    \includegraphics[width=\linewidth]{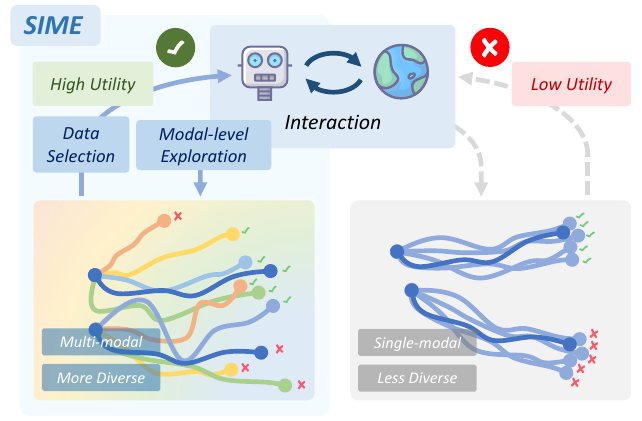}
    \caption{\textbf{Overview of SIME.} With modal-level exploration, the robot can generate more diverse and multi-modal interaction data. By learning from the most valuable trials and high-quality segments from these interactions, the robot can effectively refine its capabilities through self-improvement.}
    \label{fig:teaser}
    \vspace{-0.5cm}
\end{figure}

In this work, as shown in Fig.~\ref{fig:teaser}, we propose an approach named \textit{Policy Self-Improvement with Modal-level Exploration~(SIME)} that allows imitation learning policies to autonomously refine their capabilities through self-collected experiences, without requiring additional human intervention. Previous works~\cite{mirchandani2024so} and our experiments demonstrate that naively fine-tuning policies using self-collected interaction data is often ineffective. This is because imitation learning policies often produce repetitive or deterministic behaviors, resulting in limited diversity in the collected data. Rather than modifying the policy architecture or training objective to enhance output diversity, our goal is to develop a plug-and-play solution that can be seamlessly integrated into policy inference, without interfering with the current training process. 
Our key finding in this paper is that introducing modal-level exploration during interaction can significantly increase the diversity of the self-collected data and improve self-improvement effectiveness. Additionally, we show that reasonable inter-demo selection and intra-demo selection strategies can further enhance the sample efficiency and performance of self-improvement. By creating and focusing on the most informative self-collected data, our approach can effectively enable imitation learning algorithms to achieve self-improvement from past interactions.

To evaluate the effectiveness of \textit{SIME}, we conducted both qualitative and quantitative experiments to verify the significant improvement in the diversity of the data generated by the interaction through modal-level exploration. The experiments show that performing only one round of self-improvement on the policy using \textit{SIME} on the RoboMimic~\cite{mandlekar2021matters} benchmark results in an average improvement of approximately 16.1\%, which is roughly twice the performance of the baseline method. Furthermore, after multiple rounds of iteration, our method continues to achieve further improvements, which demonstrates the significant impact of our method on policy self-improvement.

\section{Related Works}

In this section, we review related literature on robot policy self-improvement, including reinforcement learning, residual policy learning, human-in-the-loop learning, and autonomous imitation learning.

\subsection{Reinforcement Learning}
Reinforcement learning (RL)~\cite{haarnoja2018soft,lillicrap2015continuous,schulman2017proximal} is a popular approach to enabling robots to learn from their own experiences. 
Despite their success in various domains, these methods often require extensive interactions with the environment, as well as delicate reward engineering, to achieve satisfactory performance, which limits their scalability and practicality in real-world robot manipulation settings. 
In the past decade, researchers have explored various alternatives to enhance the sample efficiency of RL algorithms. One promising direction is offline reinforcement learning~\cite{fujimoto2019off,kostrikov2021offline,kumar2020conservative}, which aims to learn policies from a fixed dataset without requiring online interactions. Compared to simple behavior cloning algorithms~\cite{mandlekar2021matters}, offline RL methods are designed to leverage extra learning signals, such as rewards and failures, to outperform the original behavior policies. However, due to the lack of online interactions and instant feedback, offline RL methods often suffer from distributional shift issues and extrapolation errors~\cite{fujimoto2019off}, which lead to conservative policy updates~\cite{kumar2020conservative} and limited performance improvements~\cite{mandlekar2021matters}.

\subsection{Residual Policy Learning}

Recently, a line of research~\cite{ankile2024imitation,haldar2023teach,lv2023sam,yuan2024policy} uses residual policy learning~\cite{silver2018residual} to bridge the gap between imitation learning and reinforcement learning. Among them, FISH~\cite{haldar2023teach} trains a parametric residual policy with RL to refine a non-parametric base policy initiated from human demonstrations, thus enabling interactive learning in the real world. ResiP~\cite{ankile2024imitation} applies residual policy learning to imitation-learning-based diffusion policies in simulation and distills the refined policy into a single vision-based model for sim-to-real transfer. Policy Decorator~\cite{yuan2024policy} shows that large imitation learning models can be effectively refined with a small RL-based model-agnostic residual policy. Though these works demonstrate that the RL-based residual policy can improve the performance of the imitation-learning-based base policies, they still require complex reward designs or suffer from sim-to-real gaps.

\subsection{Human-in-the-Loop Learning}

Another line of research~\cite{jiang2024transic,liu2022robot,luo2024precise,luo2024human} focuses on enabling robots to learn from human guidance directly, which can be more informative and efficient than sparse reward signals used in traditional real-world RL. For example, HIL-SERL~\cite{luo2024precise} introduces human corrections to SERL~\cite{luo2024serl} training, reducing the difficulty of learning challenging tasks from scratch. Sirius~\cite{liu2022robot} allows human intervention during deployment, uses weighted behavior cloning to refine the policy with the collected intervention data, and iteratively improves the performance through rounds of deployment and refinement. TRANSIC~\cite{jiang2024transic} performs human-in-the-loop data collection and trains a residual policy in addition to the base policy distilled from simulations, thus lowering the required human effort and facilitating the sim-to-real transfer. In contrast to these methods, our approach aims to enable the policy to autonomously refine its capabilities through self-driven exploration, without requiring additional human intervention.

\subsection{Autonomous Imitation Learning}

Perhaps the most closely related work to ours is autonomous imitation learning~\cite{bousmalis2023robocat,mirchandani2024so}. RoboCat~\cite{bousmalis2023robocat} proposes a self-improvement framework that enables robots to autonomously collect data and refine their policies. The robot first learns a policy from human demonstrations and other relative tasks, then uses the learned policy to practice on the new task to collect interaction data, and finally trains a new version of the policy using the collected data combined with the original demonstrations. Another recent work~\cite{mirchandani2024so} identifies that autonomously collected data is often redundant, requiring certain filtering strategies, and proposes to use state novelty as a metric to select valuable data for policy refinement. Our work extends these ideas by introducing a novel exploration approach to increase data diversity and proposing inter-demo and intra-demo selections to filter out redundant data.

\section{Preliminary}

In this section, we will introduce the background knowledge related to imitation learning and diffusion policy~\cite{chi2024diffusionpolicy}.

Given a set of expert demonstrations $\mathcal{D} = \{\mathbf{\tau}_0, \mathbf{\tau}_1, \dots, \mathbf{\tau}_N\}$, where $\mathbf{\tau}_i = \{(\mathbf{o}_t^i, \mathbf{a}_t^i)\}_{t=1}^{T_i}$ is a trajectory of observation-action pairs, the goal of imitation learning~\cite{chi2024diffusionpolicy, florence2022implicit, mandlekar2021matters,shafiullah2022behavior, wang2024rise} is to learn a policy $\pi_{\theta}:\mathcal{O} \rightarrow \mathcal{A}$ that maps raw observations to robot actions. The policy, parameterized by $\theta$, is optimized to minimize the discrepancy between the expert demonstrations and the policy's behavior. 

One recent advance in imitation learning is the diffusion policy~\cite{chi2024diffusionpolicy}, which models the robot's visuomotor behavior as a conditional denoising diffusion process. Specifically, the diffusion policy learns to generate noise-free action sequences $\mathbf{a}_t^0, \mathbf{a}_{t+1}^0, \dots, \mathbf{a}_{t+H-1}^0$ by iteratively refining noisy actions through a series of denoising steps, which are formulated as:
\begin{gather}
\mathbf{a}_{t:t+H}^{K}\sim\mathcal{N}(0,\sigma_K^2I),\\
\mathbf{a}_{t:t+H}^{k-1} = \alpha(\mathbf{a}_{t:t+H}^k-\gamma\epsilon_\theta(\mathbf{a}_{t:t+H}^k,\mathbf{o}_t, k)+\mathcal{N}(0,\sigma_k^2I)),
\end{gather}
where $\mathbf{a}_{t:t+H}^{k}$ denotes the action sequence starting from timestep $t$ at noise level $k$ with horizon $H$, $\mathbf{o}_t$ is the observation, and $\epsilon_\theta$ is a neural network that predicts the noise at each denoising step. The policy can be trained by minimizing the difference between the predicted and true noise, namely:
\begin{equation}
\mathcal{L}(\theta) = \mathbb{E}_{\mathbf{o}, \mathbf{a} \sim \mathcal{D}, \epsilon \sim \mathcal{N}(0, \sigma^2I)} \left[ \|\epsilon^k - \epsilon_\theta(\mathbf{a}+\epsilon^k,\mathbf{o}, k)\|^2 \right].
\end{equation}

Due to its effectiveness in modeling visuomotor behavior, we adopt the diffusion policy as our base policy. 

\section{Method}

\subsection{Policy Self-improvement}

In this paper, we explore how to achieve effective policy self-improvement.
The learned policy  $\pi_\theta$ often fails to achieve optimal performance due to limitations in the dataset $\mathcal{D}$’s scale or diversity. The goal of policy self-improvement is to enable policy to generate a new self-collected dataset $\mathcal{D^\prime}$ through interaction with the environment, and use these data $\mathcal{D}\cup\mathcal{D}^\prime$ to fine-tune the policy, thereby enhancing its performance.

One issue with fine-tuning the policy using self-collected data is that the learned policy may easily collapse to single-modal solutions, overfitting to the dataset and failing to generalize to new scenarios. In such cases, the policy tends to generate data similar to the original dataset in scenarios where it is already proficient, but struggles to create effective successful trajectories in scenarios where it is less capable. Fine-tuning the policy from new data generated through such interactions is unlikely to lead to improvements in the policy, as it mainly reinforces existing behaviors rather than introducing new, diverse behaviors.

Previous works tend to infuse noise into the actions during policy inference, in the hope of encouraging exploration. This is typically done by sampling actions from a Gaussian mixture model with enlarged variance, or by directly adding noise to the generated actions at each timestep. These action-level exploration methods often only make the output actions more random, without increasing the modalities of the interactions. For diffusion policy, 
though it can produce a more diverse action distribution by performing stochastic Langevin dynamics sampling on the gradient field of the action score function~\cite{chi2024diffusionpolicy}, past research~\cite{ankile2024imitation} and our experiment in Sec.~\ref{sec:diverse} find that it still struggles with multi-modal exploration, even when the random level of the denoising process is set to its maximum ($\eta=1$). The diffusion policy tends to output action sequences with low diversity or even deterministic behaviors. This becomes a significant issue in self-improvement scenarios, where the policy needs to explore a wide range of actions to refine its capabilities.

\subsection{Modal-level Exploration}

\begin{figure}[t]
    \centering
    \includegraphics[width=\linewidth]{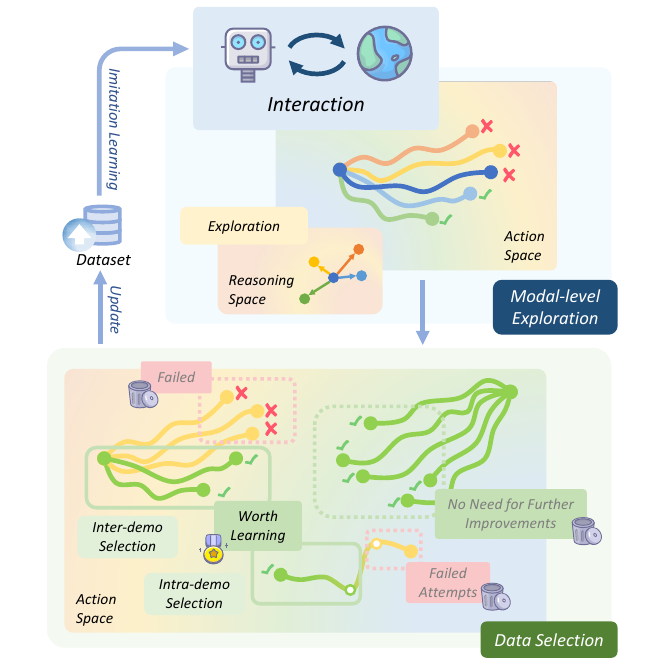}
    \caption{\textbf{Pipeline overview.} The robot learns a policy from human demonstrations, explores multi-modal interaction behaviors in the reasoning space, collects and selects valuable trajectories and segments, and refines the policy.}
    \label{fig:pipeline}
    \vspace{-0.3cm}
\end{figure}

To address this issue, we propose a simple yet effective approach, \textit{SIME}, to enhance the modal-level exploration capabilities of the diffusion policy during inference. Our key idea is to introduce a modal modulation factor $\gamma$ that allows us to inject noise into the model’s \textit{reasoning} space, rather than merely adding noise to the \textit{action} space, where
\begin{gather}
    \epsilon_t^k = \epsilon_\theta(\mathbf{a}_{t:t+H}^k, o_t, \gamma,  k).
\end{gather}
This enables more diverse and meaningful exploration, enhancing the policy’s ability to generate varied interactions without resorting to simply randomizing the actions.

Specifically, since our goal is to develop a plug-and-play method that can be seamlessly applied to existing policies, we preserve the original diffusion policy training process, and only introduce modal-level exploration to the model's conditional latent variables $z_k$ during the inference phase. In this way, our approach avoids adding complexity to model training and enhances the ease of use of the method, which can be formulated as:
\begin{gather}
    \mathbf{n} \sim \mathcal{N}(0, \sigma^2I),\\
    z_t^k = \psi(\mathbf{o}_t) + \gamma(k) \mathbf{n},
    \\
    \epsilon_t^k = \epsilon_\theta(\mathbf{a}_{t:t+H}^k, z_t^k, k),
\end{gather}
where $\psi$ is the observation encoder, and $\epsilon_t^k$ is the predicted noise at timestep $t$ and noise level $k$. By introducing the proposed perceptive perturbation, we can influence the cognitive process of the policy, driving exploratory behaviors in the reasoning space.

Additionally, inspired by~\cite{sadat2023cads}, we use a linear annealing schedule to gradually decrease the introduced noise as the denoising process proceeds:
\begin{gather}
    \gamma(k) = \left\{\begin{array}{ll}
        1 & k\leq \kappa_1,\\
        \frac{\kappa_2-k}{\kappa_2-\kappa_1} & \kappa_1 < k < \kappa_2,\\
        0 & k\geq \kappa_2,
    \end{array}\right.
\end{gather}
where $\kappa_1, \kappa_2 \in [0, K]$ are hyperparameters. This design ensures that the policy explores wider space to create diverse manipulation solutions at the beginning of the denoising process, while still maintaining the ability to generate high-precision actions in the end.

\subsection{Demo Selection}

With this modal-level exploration, the policy is designed to pursue multi-modal solutions and collect more diverse data during interaction. 
While we still need effective selection to fulfill the most valuable parts of the self-collected data. Thus we introduce inter-demo selection and intra-demo selection to further enhance the learning effectiveness and efficiency of self-improvement.

\paragraph{Inter-demo selection} aims to filter out less informative trajectories, focusing the learning process on the most valuable data. Our key insight is that \textit{accidental success in challenging scenarios} holds the highest learning value. To identify these, we conduct multiple trials from the same initial environment states. Challenging scenarios are considered as those with a success rate below a certain threshold $\theta$. In these cases, we retain only the successful portions of the trials, discarding the rest and other unchallenging cases. The retained trajectories, representing successful operations in difficult situations, are considered to have high value for learning.

\paragraph{Intra-demo selection} However, such trajectories can be redundant and noisy, particularly in difficult tasks, as they often involve multiple self-corrections following initial mistakes before reaching the final success. Intra-demo selection addresses this by enforcing the training process to focus on segments with high learning value. It allows the model to concentrate on the corrective behaviors that lead to task completion, rather than erroneous mistakes. This selection can be done through manual annotation. In our experiments, we also explore automating this selection procedure using Implicit Q-Learning (IQL)~\cite{kostrikov2021offline}. 

Specifically, we modify the approach to estimate the segment-wise value increments and train the policy using weighted regression~\cite{peng2019advantage,wang2018exponentially}:

\begin{align}
    \mathcal{L}(\theta) &= \mathbb{E}_{\mathbf{o_t}, \mathbf{a}_{t:t+H} \sim \mathcal{D}, \epsilon \sim \mathcal{N}(0, \sigma^2I)} \nonumber\\
    &\left[ \|\epsilon^k - \epsilon_\theta(\mathbf{a}_{t:t+H}+\epsilon^k,\mathbf{o}_t, k)\|^2 \exp\left(\frac{1}{\beta}(V_{t+H}-V_t)\right)\right].\nonumber\\
\end{align}

Leveraging the proposed pipeline, we enable the policy to generate diverse, multi-modal data during interactions and efficiently utilize this data for self-improvement.

\section{Experimental Result}

In this section, we introduce the implementation details and then we address the following questions through experiments:
\begin{itemize}
    \item Can modal-level exploration enhance the policy’s exploration capabilities and generate diverse interaction data?
    \item Is the data generated through \textit{SIME} more effective for policy self-improvement?
    \item Can inter-demo selection and intra-demo selection further improve data utilization efficiency?
    \item Are the design choices in \textit{SIME} effective in improving the overall process?
\end{itemize}

\subsection{Implementation Details}

\subsubsection{Benchmark}
We evaluate \textit{SIME} on 5 tasks from RoboMimic~\cite{mandlekar2021matters}, a benchmark suite for robot learning from demonstrations. Those tasks include: \textit{Lift}, \textit{Can}, \textit{Square}, \textit{Transport}, and \textit{ToolHang}, each of which requires the robot to perform a specific manipulation task in a simulated environment. All tasks include data from both the state-space and image-space observation domains. For more details, please refer to~\cite{mandlekar2021matters}.

\subsubsection{Policy}
We use Diffusion Policy~\cite{chi2024diffusionpolicy} as the base policy and ResNet-18~\cite{he2016deep} as the image encoder for image-space experiments. To accelerate the inference speed, we use a DDIM~\cite{song2020denoising} scheduler with 100 training iterations and 20 inference iterations. We train most tasks using a batch size of 64 and a learning rate of 3e-4 for 1000 epochs.

Given that the diffusion policy already achieves near 100\% success rate on most tasks with 200 proficient human (PH) demonstrations~\cite{chi2024diffusionpolicy}, we focus on a more challenging few-shot setting: to learn the policy from a limited data source and refine it through self-improvement. In our experiments, we use 20 trajectories for task \textit{Can}, \textit{Square}, \textit{Transport}, 10 trajectories for \textit{Lift} since it is relatively easier, and 40 trajectories for \textit{ToolHang} since it requires more precise and long-horizon manipulation. This also aligns with our expectations for self-improvement: by relying on fewer human-provided data and learning more from its own interaction experiences, the system can reduce the cost of data.

To lower the variance of the results and obtain reliable experiment results, we repeated each experiment 4 times with different random seeds. The mean and standard deviation of the success rate are reported based on the results of the 4 runs. During the training, we use the same hyperparameters for the same task. Since different ways of data collection may result in different data scales, we fix the number of training iterations to ensure a fair comparison, which means that each experiment was conducted with the same computational resources. The last checkpoint of each training run is used for evaluation. During the evaluation, 100 random initial states are sampled for each task, and we offer the policy 5 attempts to complete the task from each initial state. The final success rate is calculated as the percentage of successful attempts, based on $4\times 100\times 5=2000$ total trials for each. 

\subsection{Diversity Analysis on Self-Collected Data}\label{sec:diverse}

\begin{figure}[h]
 \centering
 \includegraphics[width=0.75\linewidth ]{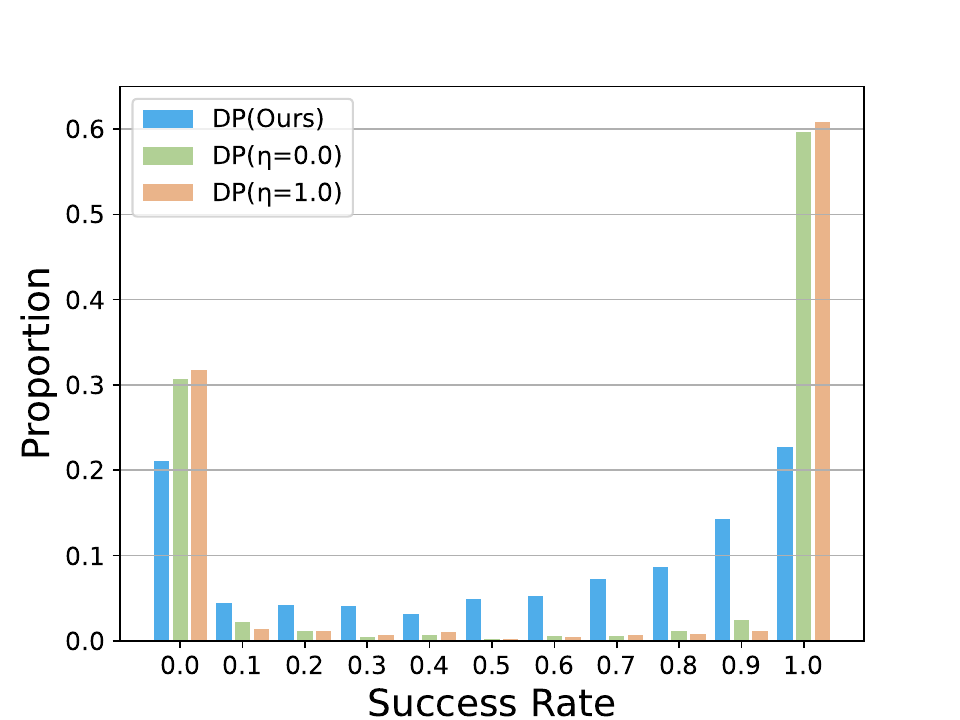}
  \caption{\textbf{Diversity analysis.} We tested 1,000 different scenarios, conducting 10 trials for each. After independently calculating the success rate for each scenario, we analyzed the distribution of success rates. As shown, modal-level exploration allows the policy to exhibit diverse behaviors.} \label{fig:succdis}
\end{figure}

We first analyze the diversity of the self-collected trajectories generated by the diffusion policy and the impact of adding modal-level exploration on its diversity. 
To do so, we selected 1,000 different initial states for task \textit{Can} and conducted 10 trials for each state with the policy trained on 20 demonstrations. We analyze the distribution of success rates across multiple attempts from the same initial state. The results are presented in Fig.~\ref{fig:succdis}.

From the figure, we can observe that the diffusion policy without modal-level exploration results in either complete success or complete failure in 10 trials in approximately 91\% of the cases. This suggests that, in the same scenarios, the policy tends to follow the same approach to complete the task, lacking diversity. Even by adjusting $\eta$~(for the detailed definition of $\eta$, please refer to~\cite{song2020denoising}) to alter the randomness of the diffusion policy, we are unable to effectively increase the diversity of the output trajectories. However, after introducing modal-level exploration, we observe a significant change: only about 44\% of the scenarios result in either complete success or complete failure in 10 trails under the same initial state, while 56\% of the scenarios exhibit partial success and partial failure due to the multi-modal attempts of the policy. This indicates an increase in diversity.

In addition, the policy without modal-level exploration fails to produce any successful attempts in approximately 32\% of the scenarios. While introducing modal-level exploration results in an 8\% decrease in the average success rate across all attempts, the percentage of scenarios with no successful attempts at all drops to 21\%. In other words, the use of modal-level exploration leads to an 11\% improvement in the success rate at the scenario level. This suggests that our exploration method allows the policy to discover solutions in cases that previously failed.

\begin{figure}[t]
 \centering
 \includegraphics[width=1.0\linewidth]{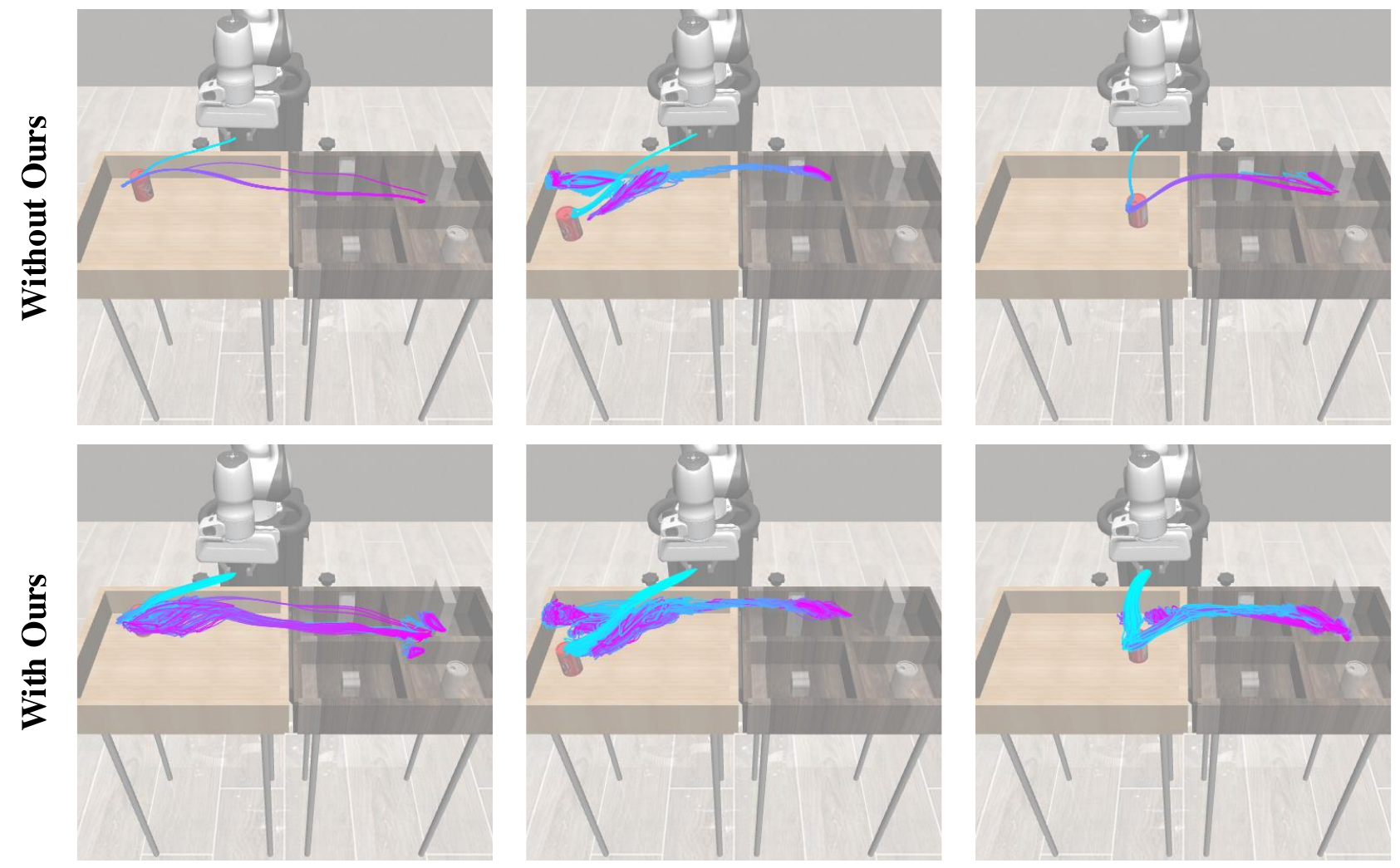}
  \caption{\textbf{Trajectory comparison.} By conducting 100 tests with the same initial state, we observe that the diffusion policy’s output lacks diversity. However, introducing modal-level selection significantly increases the diversity of the generated behaviors.} \label{fig:vistraj}
\end{figure}

Additionally, we have visualized the changes in the trajectories generated by the diffusion policy before and after the introduction of modal-level exploration. As shown in Fig.~\ref{fig:vistraj}, before introducing modal-level exploration, the policy tends to follow the same trajectory to complete the task. After introducing modal-level exploration, a clear multi-modal behavior emerges, which leads to an increase in the diversity of the generated interaction data.

In summary, the introduction of modal-level exploration enables the policy to generate more diverse interaction experiences. Particularly for scenarios where the policy would completely fail, modal-level exploration helps guide the policy to explore successful attempts in these previously challenging scenarios, making the generated data highly valuable for learning. Next, we analyze the impact of the data generated after the introduction of modal-level exploration on the policy’s self-improvement.

\subsection{Performance of SIME}

To start our experiment, we first train a diffusion policy $\pi_{0}$ using limited expert demonstrations $\mathcal{D}_{init}$. We then evaluate the policy $\pi_{0}$ without any further modification to record the initial performance. The generated trajectories $\mathcal{D}_{0}$ are saved and used as the self-collected data for baselines. Next, we apply \textit{SIME} to the initial policy and conduct evaluation once again to collect trajectories $\mathcal{D}^\prime_{0}$. The same data filtering and selection strategies $\mathcal{F}$ are applied to both datasets to ensure a fair comparison. To balance effectiveness and efficiency, we set the success threshold $\theta$ for inter-demo selection to 0.5. We then train two new policies $\pi_{1}$ and $\pi_{1}^\prime$ using $\mathcal{D}_{init}\cup\mathcal{F}(\mathcal{D}_{0})$ and $\mathcal{D}_{init}\cup\mathcal{F}(\mathcal{D}^\prime_{0})$ respectively, and evaluate their performance. The results are presented in Table~\ref{tab:state} and Table~\ref{tab:image}.

\begin{table*}[htbp]
    \centering
    \caption{Experiment Results on State-based Benchmark}
    \label{tab:state}
    \begin{tabular}{c|c|ccccc|c}
    \toprule
    \multicolumn{2}{c|}{}
    &Lift-10&Can-20&Square-20&Transport-20&ToolHang-40&Average Increment\\
    \midrule
    \multicolumn{2}{c|}{Before Self-Improvement $\pi_{0}$}
    &$0.781\pm0.009$&$0.658\pm0.025$&$0.534\pm0.016$&$0.555\pm0.040$&$0.112\pm0.011$&$-$\\
    \midrule
    \multirow{2}{*}{After Self-Improvement}
    &Baseline $\pi_{1}$&$0.797\pm0.026$&$0.688\pm0.036$&$0.556\pm0.026$&$0.557\pm0.011$&$0.190\pm0.028$&$+5.6\%$\\
    &Ours $\pi^\prime_{1}$&$0.848\pm0.023$&$0.729\pm0.028$&$0.620\pm0.009$&$0.585\pm0.019$&$0.183\pm0.020$&$+12.3\%$\\
    \bottomrule
    \end{tabular}
\end{table*}

\begin{table*}[htbp]
    \centering
    \caption{Experiment Results on Image-based Benchmark}
    \label{tab:image}
    \begin{tabular}{c|c|ccccc|c}
    \toprule
    \multicolumn{2}{c|}{}
    &Lift-10&Can-20&Square-20&Transport-20&ToolHang-40&Average Increment\\
    \midrule
    \multicolumn{2}{c|}{Before Self-Improvement $\pi_{0}$}
    &$0.901\pm0.033$&$0.548\pm0.039$&$0.244\pm0.025$&$0.445\pm0.054$&$0.144\pm0.036$&$-$\\
    \midrule
    \multirow{2}{*}{After Self-Improvement}
    &Baseline $\pi_{1}$&$0.876\pm0.030$&$0.585\pm0.017$&$0.306\pm0.036$&$0.569\pm0.035$&$0.205\pm0.049$&$+11.3\%$\\
    &Ours $\pi^\prime_{1}$&$0.914\pm0.016$&$0.651\pm0.050$&$0.371\pm0.021$&$0.593\pm0.017$&$0.207\pm0.047$&$+19.9\%$\\
    \bottomrule
    \end{tabular} 
\end{table*}

From the tables, we can observe that our introduction of self-improvement leads to policies that outperform initial policies across all tasks, leading to an average relative increment of 12.3\% and 19.9\% for state-based tasks and image-based tasks, respectively. Compared with baselines, the proposed \textit{SIME} achieves superior performance improvements, with an additional 8.6\% improvement in vision space and 6.9\% in state space. This indicates the effectiveness of our modal-level exploration approach and self-improvement pipeline in enhancing the policy's performance. It is also important to note that both the baseline method and our approach utilize inter-demo selection in this experiment, which, as demonstrated in Sec.~\ref{sec:selection}, has a significant impact on self-improvement. A large portion of the improvements observed in the baseline method also stem from this. The impact of intra-demo selection will be discussed in Sec.~\ref{sec:selection} later.

\begin{figure}[h]
 \centering \includegraphics[width=0.75\linewidth ]{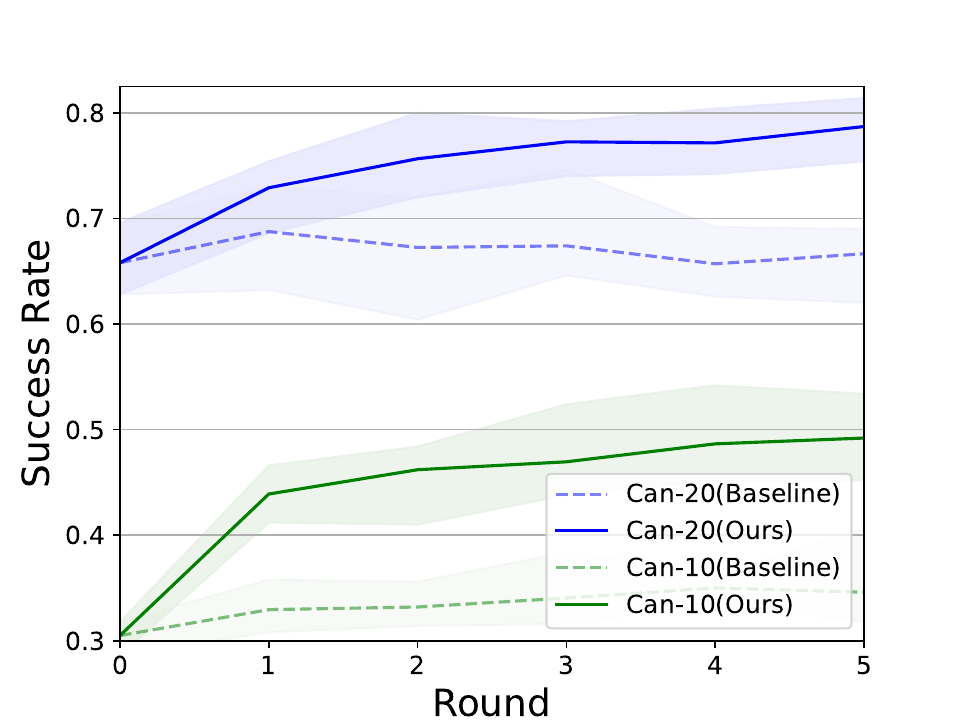}
  \caption{\textbf{Multi-round self-improvement results} comparing the baseline method and our approach on the \textit{Can} task, starting with 10 and 20 human demonstrations.} \label{fig:multiround}
\end{figure}

To investigate whether the policy can be further improved by continuing the self-improvement process, we also conducted a multi-round experiment. After finishing the self-improvement process mentioned above, we collect new trajectories generated by the refined policies and follow the same procedure to self-improve the policies for 5 rounds. The result is shown in Fig.~\ref{fig:multiround}.

\begin{table}[h]
    \centering
    \caption{Evaluation on Demo Selection}\label{tab:selection}
    \begin{tabular}{ccccc}
    \toprule
    \makecell[c]{Inter-Demo\\Selection}& \makecell[c]{Intra-Demo\\Selection}& \makecell[c]{Demo\\Nums} & Result & Increment\\
    \midrule
    &&310&$0.708\pm0.025$&$+7.6\%$\\
    Random&&40&$0.672\pm0.020$&$+2.1\%$\\
    SR<0.5&&40&$0.729\pm0.028$&$+10.8\%$\\
    SR<0.5&\checkmark&40&$0.734\pm0.028$&$+11.6\%$ \\
    \bottomrule
    \end{tabular}
\end{table}

From the figure, we can observe that our proposed \textit{SIME} continuously improves the policy’s performance over multiple iterations. In contrast, the baseline method, although showing some improvement in the first iteration, experiences diminishing gains in subsequent rounds, and even sees its improvements vanish. This demonstrates that as the number of iterations increases, the advantage of \textit{SIME} over the baseline method continues to grow.

\subsection{Evaluation on Data Selection}\label{sec:selection}

We also conducted experiments to evaluate the impact of inter-demo and intra-demo selection strategy on \textit{Can-20} self-improvement. The result is reported in Tab.~\ref{tab:selection}.

\paragraph{Inter-demo Selection}
To evaluate the impact of inter-demo selection on \textit{SIME}, we designed two additional experiments. First, we removed the inter-demo selection approach to the data collected during interactions and used the raw data for model improvement. This resulted in a drop in the performance improvement from 10.8\% to 7.6\%. We noticed that in this experiment, the scale of the data used for training changed significantly. To further isolate the impact of data scale on learning, we set up a control experiment where we randomly selected trajectories from the raw data for training, while ensuring that the number of samples remained consistent across different inter-demo selection strategies. Under such condition, the performance improvement from \textit{SIME} almost vanished. This demonstrates that our proposed inter-demo selection approach can significantly enhance the effectiveness and sample efficiency of \textit{SIME}.

\paragraph{Intra-demo Selection}
We also validated the effect of intra-demo selection through experiments. As shown in the table, introducing intra-demo selection resulted in an additional performance improvement.

In experiments involving demo selection, our straightforward yet effective heuristic approach demonstrated significant performance improvements. These findings not only validate the practical value of intuitive selection strategies, but more importantly, highlight their potential to inspire the community in data selection for policy self-improvement. Building upon this foundation, we will focus on developing more effective data selection in future work.

\subsection{Ablation Study}

The following experiments in Tab.~\ref{tab:ablation} validate the effectiveness of several key components in \textit{SIME} on the \textit{Can-20} benchmark.

Directly adding noise to the action space for exploration does not result in data that positively impacts the model’s self-improvement. Additionally, adjusting the randomness of the diffusion process by changing $\eta$ to $1.0$ does not lead to significant improvements in self-improvement either. However, the annealing strategy we introduced in this paper has a positive effect on the performance of \textit{SIME}.

\begin{table}[h]
    \centering
    \caption{Ablation Study}\label{tab:ablation}
    \begin{tabular}{cccc}
    \toprule
    & \makecell[c]{Demo\\Nums} & Result & Increment\\
    \midrule
    Action-level Noise&69&$0.591\pm0.048$&$-10.2\%$\\
    Diffusion Randomness&26&$0.667\pm0.024$&$+1.4\%$\\
    Ours w/o Annealing&37&$0.724\pm0.033$&$+10.0\%$\\
    Ours&40&$0.729\pm0.028$&$+10.8\%$\\
    \bottomrule
    \end{tabular}
\end{table}

\begin{table}[h]
    \centering
    \caption{Experiment Results in the Real World}\label{tab:realworld}
    \begin{tabular}{cccc}
    \toprule
    & \makecell[c]{Demo\\Nums} & Result & Increment\\
    \midrule
    Initial $\pi_0$&50&$0.34$&$-$\\
    Baseline $\pi_1$&67&$0.48$&$+41.2\%$\\
    Ours $\pi^\prime_1$&67&$0.74$&$+117.6\%$\\
    \bottomrule
    \end{tabular}
\end{table}

\subsection{Real-World Experiments}

We also demonstrate the performance of \textit{SIME} in real-world experiments. To set up the real-world experiments, we use a Flexiv Rizon4\footnote{https://www.flexiv.cn/product/rizon} robot arm equipped with a Robotiq 2F-85\footnote{https://robotiq.com/products/adaptive-grippers} gripper. The gripper finger is changed to TPU soft finger\footnote{https://umi-gripper.github.io}. We set up 2 RealSense D435i\footnote{https://www.intelrealsense.com/depth-camera-d435i/} depth cameras to observe the environment, one mounted on the robot wrist and the other fixed on a side perspective. 

We conducted real-world experiments on a \textit{Cup Stacking} task, where the robot is required to place a red plastic cup inside a metal cup. At the start of each evaluation, both cups are randomly positioned on the platform. A total of 50 human-teleoperated demonstrations were provided for imitation learning. The resolution of the image observations was fixed at $480\times640$ throughout the experiment.

The same policy architecture and training hyperparameters are used as in the simulation, with batch size 64, learning rate 3e-4, and 1000 training epochs. During the evaluation, we randomly initiated 10 different states and offered the policy 5 attempts to complete the task from each initial state. The final performance metric is defined as the success rate across all attempts. We also applied the same self-improvement pipeline as in the simulation to refine the policy using self-collected data and evaluated the performance of the refined policies. The results are presented in Tab.~\ref{tab:realworld}.

\begin{figure}[t]
    \centering
    \includegraphics[width=\linewidth]{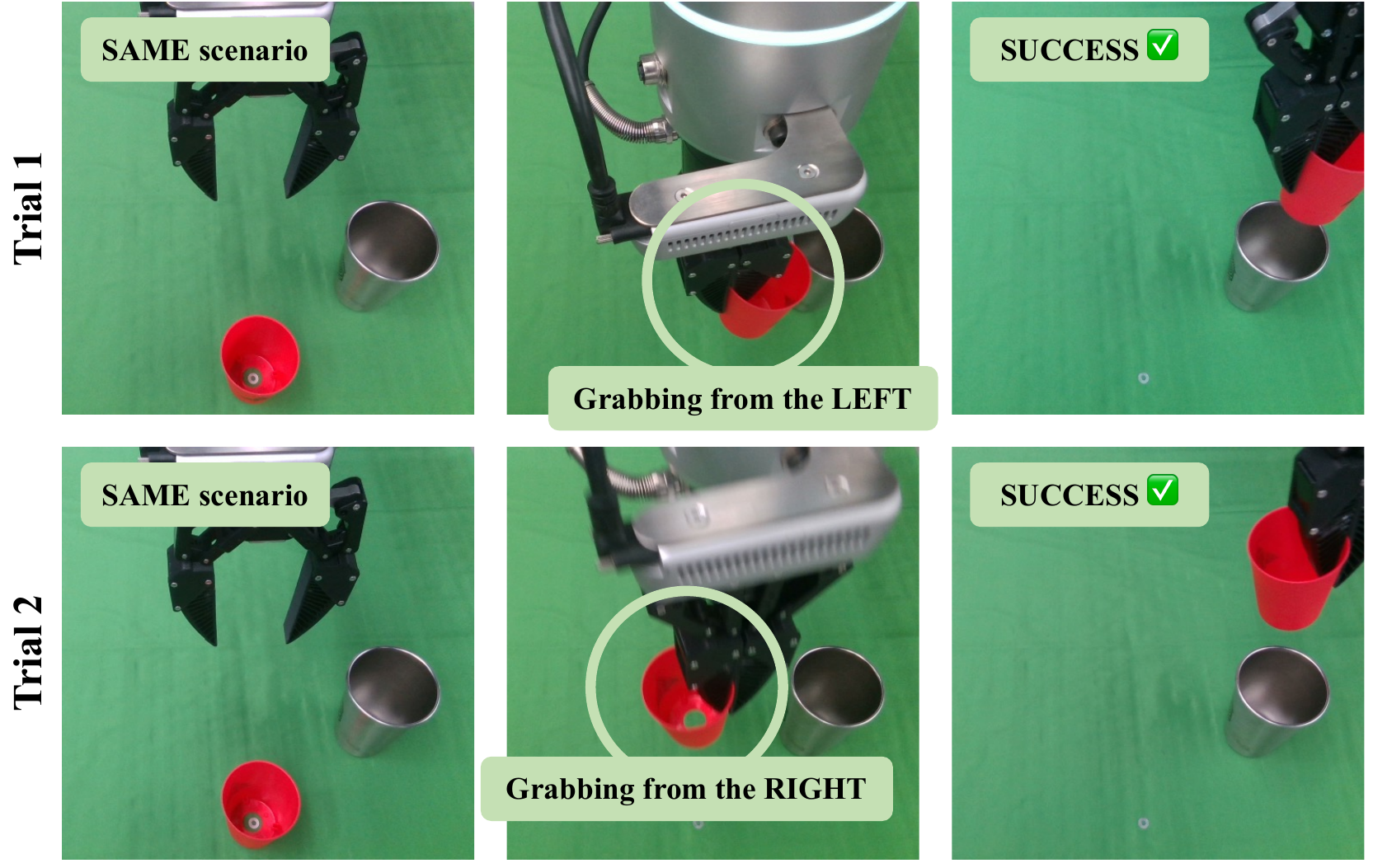}
    \caption{\textbf{Keyframes of real-world manipulation.} During two different trials, the policy tries to grab the cup in two different ways (from the left side and the right side).}
    \label{fig:realdemo}
\end{figure}

To our surprise, in real-world experiments, \textit{SIME} outperforms the baseline method by a significant margin, with an improvement of 117.6\%. We also find that without the introduced modal-level exploration, the policy tends to behave in a single-modal manner, often failing to complete the task even given multiple attempts. While with the exploration, a clear multi-modal behavior emerges. As demonstrated in Fig.~\ref{fig:realdemo}, starting from the same initial state, the robot can grab the cup from different sides. This indicates that our proposed \textit{SIME} can bootstrap the multimodality of self-collected data and improve the policy's performance in practical deployment.

\section{Conclusion}

In this paper, we present \textit{SIME}, a novel self-improvement approach that enables imitation learning policies to enhance their performance through interaction experiences. By incorporating modal-level exploration, we enable the policy to discover multi-modal solutions and gather diverse interaction data. We also propose inter-demo and intra-demo selection mechanisms to filter out less informative data, focusing on the most valuable trajectories or segments for learning. Our experiments demonstrate that \textit{SIME} significantly increases the diversity of self-collected data and improves policy performance in both simulation and real-world scenarios, with further improvements possible through multiple rounds of self-bootstrapping. Additionally, our results highlight the critical role of data selection in self-improvement. We hope this work will inspire further research and contribute to the development of sample-efficient, self-improving robot learning systems.



\printbibliography

\end{document}